# Evaluating Text Style Transfer Evaluation: Are There Any Reliable Metrics?

**Sourabrata Mukherjee[1], Atul Kr. Ojha[2], John P. McCrae[2], Ondřej Dušek[1]**

[1]Charles University, Faculty of Mathematics and Physics, Prague, Czechia
[2]Insight Research Ireland Centre for Data Analytics, DSI, University of Galway, Ireland
{mukherjee,odusek}@ufal.mff.cuni.cz
{atulkumar.ojha,john.mccrae}@insight-centre.org

## Abstract

Text style transfer (TST) is the task of transforming a text to reflect a particular style while preserving its original content. Evaluating TST outputs is a multidimensional challenge, requiring the assessment of style transfer accuracy, content preservation, and naturalness. Using human evaluation is ideal but costly, as is common in other natural language processing (NLP) tasks; however, automatic metrics for TST have not received as much attention as metrics for, e.g., machine translation or summarization. In this paper, we examine both set of existing and novel metrics from broader NLP tasks for TST evaluation, focusing on two popular subtasks—sentiment transfer and detoxification—in a multilingual context comprising English, Hindi, and Bengali. By conducting meta-evaluation through correlation with human judgments, we demonstrate the effectiveness of these metrics when used individually and in ensembles. Additionally, we investigate the potential of large language models (LLMs) as tools for TST evaluation. Our findings highlight newly applied advanced NLP metrics and LLM-based evaluations provide better insights than existing TST metrics. Our oracle ensemble approaches show even more potential.

## 1 Introduction

Text style transfer (TST) refers to the task of modifying a given text to reflect a specific style while preserving its original content (Hu et al., 2022). Previous work in this domain has explored altering various stylistic dimensions, such as sentiment (Prabhumoye et al., 2018), romantic tone (Li et al., 2018), politeness (Madaan et al., 2020), or political slant (Prabhumoye et al., 2018). Different modeling approaches have been proposed for TST, including methods that manipulate latent representations of text (Zhao et al., 2018; Prabhumoye et al., 2018) and techniques that identify and replace style-related lexicons directly (Li et al., 2018; Fu et al., 2019). Despite the growing interest in TST, reliably assessing the performance of TST models continues to be a bottleneck (Hu et al., 2022). While human evaluation is often regarded as the standard for capturing subtle cues in style, it is expensive, time-intensive, and difficult to reproduce at scale (Briakou et al., 2021b). Consequently, automated metrics have become a proxy for human judgment, but there is a notable lack of standardization and consensus on which metrics best capture style transfer accuracy, content preservation, and overall naturalness (Mir et al., 2019a; Briakou et al., 2021a). In addition, large language models (LLMs) could serve as alternatives to traditional human evaluation and automated metrics for TST evaluation (Ostheimer et al., 2024). However, the rapid evolution of LLMs, particularly for closed-source models, raises concerns about reproducibility (Gao et al., 2024; Chen et al., 2024).

We address this gap by examining existing and novel metrics for two popular TST subtasks: *sentiment transfer* (Prabhumoye et al., 2018) and *detoxification* (Dementieva et al., 2022). Our experiments span a multilingual setting, covering English, Hindi, and Bengali, to investigate the utility of these metrics across diverse linguistic contexts. We then conduct a meta-evaluation of the proposed metrics by measuring their correlation with human judgments. To further explore the potential of automated metrics, we also combine them in ensembles, experimentally creating hybrid scores. Additionally, we investigate the applicability of large language models (LLMs) as an alternative evaluation tool. Our results show that existing metrics newly applied to TST, hybrid approaches, and LLMs can improve correlation with human evaluations, offering a more robust and comprehensive assessment of TST outputs. Our experimental code and resources are released on GitHub.[1]

---

[1]https://github.com/souro/tst_evaluation

## 2 Related Work

TST tasks are traditionally evaluated using three key dimensions: *style transfer accuracy*, *content preservation*, and *fluency* (Mukherjee and Dušek, 2024; Hu et al., 2022; Jin et al., 2022). Prior work underscores the challenge of jointly capturing subtle stylistic nuances and preserving semantic content (Briakou et al., 2021b; Tikhonov et al., 2019).

**Style Transfer Accuracy** A common approach is to train a dedicated classifier to check whether the transformed text reflects the intended style (Prabhumoye et al., 2018; Shen et al., 2017). Alternatively, unsupervised methods rely on distributional shifts in style-related features (Yang et al., 2018; Tikhonov et al., 2019).

**Content Preservation** Metrics such as *BLEU* (Papineni et al., 2002) and embedding-based similarity (Rahutomo et al., 2012; Reimers and Gurevych, 2019) often serve as proxies for semantic fidelity. However, they may overlook nuances introduced by stylistic transformations in both single-language and multilingual contexts (Yamshchikov et al., 2021; Briakou et al., 2021a), and recent studies highlight the shortcomings of traditional similarity measures when evaluating paraphrase-like modifications (Yamshchikov et al., 2021; Briakou et al., 2021b).

**Fluency** *Fluency* is typically estimated using perplexity from a pre-trained language model such as *GPT-2* (Radford et al., 2019). Nonetheless, perplexity may fail to capture context-specific grammatical coherence, especially if the style domain diverges from the model's training data (Tikhonov et al., 2019; Briakou et al., 2021b), and can yield inconsistent performance across languages (Briakou et al., 2021a).

## 3 Metrics Compared

We follow the criteria of transfer accuracy, content preservation, and fluency described in Section 2, and we conduct evaluations in two scenarios: (1) *reference-based*, where metrics are computed against a reference text (when available), and (2) *reference-free*, where metrics directly compare the generated text against the source text (measuring similarity or distance from the original), without requiring a reference.

**Previously Used TST Metrics** For style transfer accuracy, we include *Sentence Accuracy* based on a fine-tuned *XLM-RoBERTa-base* (Conneau et al., 2020) classifier (Prabhumoye et al., 2018), and *WMD* (Kusner et al., 2015; Wei et al., 2023; Mir et al., 2019b). For content preservation: *BLEU* (Papineni et al., 2002; Tikhonov et al., 2019), *Cosine Similarity* (Rahutomo et al., 2012; Reimers and Gurevych, 2019), *Masked BLEU and Masked Cosine Similarity* (Mukherjee et al., 2022), *ROUGE-2* and *ROUGE-L* (Lin and Hovy, 2003; Lin, 2004; Lin and Och, 2004; Yamshchikov et al., 2021). For fluency, we use *Perplexity* of *GPT-2* (Radford et al., 2019; Briakou et al., 2021c) and *MGPT* (Shliazhko et al., 2024).

**Newly Applied Text Metrics** We expand the TST evaluation by incorporating additional metrics from related NLP tasks, categorizing them into trainable and non-trainable metrics as well as word-overlap-based and embedding-based measures.

For style transfer accuracy, we utilize non-trainable statistical measures such as *Earth Mover's Distance (EMD)* (Rubner et al., 2000), *KL Divergence* (Kullback, 1997), *Cosine Similarity* (Rahutomo et al., 2012), and *Jensen-Shannon Divergence* (Lin, 1991), which quantify the distributional shift between source and generated text. Additionally, we incorporate a trainable *Classifier Confidence* score, derived from the Sentence Accuracy classifier described earlier.

For content preservation, we include both word-overlap-based and embedding-based metrics. The word-overlap-based metrics include *PINC* (Chen and Dolan, 2011), which measures the proportion of n-grams in the generated text that do not appear in the source text (higher values indicate greater lexical divergence), *METEOR* (Banerjee and Lavie, 2005), which accounts for synonymy and stemming, and *Translation Edit Rate (TER)* (Snover et al., 2006), which evaluates the number of edits required to transform the generated text into the reference. Embedding-based measures include *Word Mover's Distance (WMD)* (Kusner et al., 2015; Wei et al., 2023), *BERTScore* (Zhang et al., 2020), *$S^3BERT$* (Opitz and Frank, 2022), and *BLEURT* (Sellam et al., 2020), all of which assess content similarity based on contextualized vector representations. Additionally, we introduce *Tree Edit Distance (TED)* (Zhang and Shasha, 1989), which measures structural similarity by computing the minimum number of tree edit operations (insertion, deletion, substitution) required to transform one syntactic tree into another. This metric is particularly useful in evaluating syntactic shifts in generated text.

For fluency evaluation, we employ language model perplexity, using *Finetuned GPT-2* and *Finetuned MGPT* trained on target styles (see finetuning details in Appendix C). Lower perplexity scores indicate higher fluency, as they reflect the model's confidence in the generated text.

**Novel Metrics** We analyze the structural similarity between the source/reference and the system-generated outputs by parsing them into abstract meaning representation (AMR) (Banarescu et al., 2013) and syntactic dependency trees (Straka and Straková, 2017). AMR provides a semantic abstraction of sentences by capturing their core meaning as directed graphs, while syntactic dependencies represent the grammatical relationships between words in tree form. To measure structural similarity, we first convert syntactic dependency trees into AMR-style structure trees, ensuring both syntactic and semantic representations are in a comparable graph format. We then compute Smatch similarity (Cai and Knight, 2013) for both AMR graphs and the syntactic trees translated to AMR-style trees. Smatch (a graph-matching metric) computes the F-score between AMR graphs by aligning their nodes and edges optimally, regardless of differences in variable naming or graph representation. A higher Smatch score, i.e., a higher AMR graph and syntactic tree similarity, indicates greater preservation of meaning and syntactic structure in the transformed text.

**LLM Prompting** Following Ostheimer et al. (2024) and Mukherjee et al. (2024b), we use LLMs as TST evaluators and extend their methods to newer LLMs, more TST tasks, and additional languages. We used GPT-4 (Achiam et al., 2023) and Llama-3.1 8B (Dubey et al., 2024) to assess the TST tasks. We employed a Likert-scale-based approach to evaluate style transfer accuracy, content preservation, and fluency. To facilitate direct comparison with *Sentence Accuracy*, we also conducted a binary evaluation for style transfer accuracy (*GPT4-bin-acc, Llama-bin-acc*). Detailed prompt instructions are provided in Appendix D.

**Hybrid** We propose two ensemble-based oracle metrics – *Hybrid-Simulation* and *Hybrid-Learned* – to show the potential of integrating multiple evaluation metrics.[2] In *Hybrid-Simulation*, we first select the top three metrics (based on correlation with human judgments) for each task and language from Tables 1 and 2. We then conduct a simulation to determine the selected metrics' relative weights by tuning them on human-labeled target data and compute their geometric average to form the final ensemble score. In *Hybrid-Learned*, we train a random forest regressor (Liaw, 2002) using all available metrics as features and human ratings as the target labels. The model assigns importance scores to each metric, and we select the top three metrics with the highest normalized importance scores. Their geometric average, weighted by these importance scores, is used to generate the ensemble result. For details on the selected metrics and their respective weights, see Tables 5 and 6 in Appendix A.

---
[2]These metrics are considered "oracle", since the approach learns optimal weights based on the target data.

**Overall Score** Following Loakman et al. (2023) and Yang and Jin (2023), we adopt the geometric mean of style transfer accuracy, content preservation, and fluency as a single aggregated score for comparison. We again aim to show the potential of this approach by producing oracle metrics. Based on the Pearson correlation results from our experiments (Tables 1, 2 and, 3), we first select the best-performing metrics for these three dimensions from previously used methods (*Existing*). We also do the same selection using newly proposed methods (excluding hybrid approaches), creating the $Ours_1$ score. We then extend $Ours_1$ by incorporating the top-performing metrics from our proposed approaches, including hybrids, to construct $Ours_2$. In addition to geometric mean scores, we directly prompt *GPT-4* and *Llama* for this task. Table 7 in Appendix A detail the metrics selected for each language and task.

## 4 Experiment Setup

**Evaluation Data: Tasks, Languages and Model Outputs** We evaluate our methods on the outputs of TST models and human annotations provided by Mukherjee et al. (2024b). This comprises two TST tasks – sentiment transfer (positive to negative statements and vice versa), where data is available for English, Hindi and Bengali, and detoxification (toxic to clean text), with English and Hindi data. Model outputs for all tasks were produced by GPT-3.5 (OpenAI, 2023), LLaMA-2-7B-Chat (Touvron et al., 2023) and Mistral-7B-Instruct (Jiang et al., 2023), as well as previous finetuned BART models by Mukherjee et al. (2024a, 2023).

**Meta-Evaluation Approach** We follow common practice for meta-evaluation (Kilickaya et al.,

2017; Zhang et al., 2020; Liu et al., 2023) and compute all metrics' correlation with human judgment using Pearson (PC), Spearman (SC), and Kendall's Tau (KC) Correlations (Schober et al., 2018; Puka, 2011).

## 5 Results Analysis

Since we found that reference-based metrics generally underperform their reference-free variants, we focus on the reference-free setting in the analysis. We include reference-based results in Appendix B.

### 5.1 Style Transfer Accuracy

The results for style transfer accuracy in the reference-free setting are shown in Table 1.
**Previously Used:** *Sentence Accuracy* generally achieves moderate to good correlation with human judgments, suggesting that direct style classification accuracy can be a reliable indicator of style transfer quality. Meanwhile, *EMD* demonstrates a moderate degree of alignment, implying that capturing distributional shifts of stylistic cues correlates moderately with human perceptions.
**Newly Applied:** *Classifier Confidence*, *Cosine Similarity*, *KL Divergence*, and *Jensen-Shannon Divergence* generally exhibit stronger alignment with human judgments compared to existing metrics, highlighting the effectiveness of distributional measures for style intensity comparisons.
**LLMs:** *GPT-4* exhibits consistently high correlations, whereas *Llama* performs notably worse, although a binarized version (*Llama-bin-acc*) shows some moderate improvements.
**Hybrid:** *Hybrid-Simulation* demonstrates strong alignment with human ratings by combining multiple signals into a single score, while *Hybrid-Learned* performs comparably, though it may fall marginally below its simulation-based counterpart in certain cases.

Direct classification metrics reliably capture style accuracy, while distribution-based and LLM-based evaluations enhance overall alignment with human judgments, especially when integrated in hybrid frameworks. In English tasks, approaches like GPT-4 and hybrid methods achieve particularly high correlations, whereas in Hindi and Bengali, top metrics (e.g., KL, JS Divergence, and hybrid approaches) remain strong but show more pronounced performance gaps, potentially due to greater linguistic complexity.

### 5.2 Content Preservation

We present the meta-evaluation of content preservation metrics in a reference-free setting in Table 2.
**Previously Used:** *BLEU* generally shows low alignment with human judgments, while *Cosine Similarity* exhibits better performance in several tasks. *Masked BLEU* and *Masked Cosine Similarity* offer slight improvements over their unmasked counterparts, yet they still lag behind more recent methods. *ROUGE-2* and *ROUGE-L* provide moderate correlations but do not consistently outperform newer metrics.
**Newly Applied:** *BLEURT* remains consistently reliable, while *BERTScore* also proves robust across various styles and languages. *TER* and *TED* offer competitive results, particularly for certain language-specific tasks. In contrast, PINC shows weak correlations, indicating its limited effectiveness in capturing content preservation.
**Novel:** *Smatch (Dependency Trees)* and *Smatch (AMR)* outperform or at least match the performance of traditional metrics, though they generally fall behind the newly introduced text-based methods and LLM-driven approaches on average.
**LLMs:** *GPT-4* achieves higher correlations than traditional metrics across different styles and languages, demonstrating its strong ability to capture human-like judgments of text transformations. In contrast, *Llama* tends to underperform, indicating considerable variability in how well different LLMs reflect stylistic and content-based shifts.
**Hybrid:** *Hybrid-Simulation* achieves robust alignment with human ratings by unifying multiple signals into a single score, whereas *Hybrid-Learned* shows comparable performance, albeit slightly trailing the simulation-based approach in some scenarios.

### 5.3 Fluency

Table 3 presents fluency evaluation results. *GPT-2 Perplexity* displays limited correlations with human judgments, while *Finetuned GPT-2 Perplexity* yields only marginal gains. *MGPT Perplexity* and *Finetuned MGPT Perplexity* provide moderate improvements under fine-tuning, underscoring the importance of multilingual modeling and style-specific training for better alignment with human fluency assessments. *GPT-4* demonstrates relatively strong correlations with human assessments of fluency for sentiment-related tasks, suggesting it captures fluidity and coherence more effectively

|  | Sentiment Transfer | | | | | | | | | Detoxification | | | | | |
|---|---|---|---|---|---|---|---|---|---|---|---|---|---|---|---|
|  | English | | | Hindi | | | Bengali | | | English | | | Hindi | | |
| Metrics | PC | SC | KC | PC | SC | KC | PC | SC | KC | PC | SC | KC | PC | SC | KC |
| *Previously used & LLMs* | | | | | | | | | | | | | | | |
| Sentence Accuracy | **0.51** | **0.49** | **0.48** | 0.61 | 0.61 | 0.59 | 0.57 | 0.57 | 0.54 | 0.36 | 0.36 | 0.35 | 0.38 | 0.37 | 0.36 |
| EMD | 0.27 | 0.24 | 0.20 | 0.36 | 0.43 | 0.34 | 0.50 | 0.52 | 0.40 | 0.29 | 0.21 | 0.17 | **0.47** | **0.53** | **0.43** |
| GPT4 | **0.92** | **0.81** | **0.79** | **0.87** | **0.84** | 0.79 | **0.82** | **0.83** | **0.77** | **0.74** | **0.72** | **0.65** | **0.74** | **0.74** | **0.68** |
| GPT4-bin-acc | 0.89 | 0.78 | 0.77 | 0.84 | 0.83 | **0.80** | 0.77 | 0.78 | 0.74 | 0.61 | 0.61 | 0.59 | 0.60 | 0.61 | 0.59 |
| Llama | 0.16 | 0.17 | 0.15 | -0.11 | -0.10 | -0.09 | -0.17 | -0.15 | -0.13 | 0.20 | 0.18 | 0.17 | 0.20 | 0.16 | 0.15 |
| Llama-bin-acc | 0.49 | 0.44 | 0.43 | 0.50 | 0.51 | 0.49 | 0.31 | 0.31 | 0.30 | 0.24 | 0.24 | 0.23 | 0.27 | 0.27 | 0.27 |
| *Newly applied & Novel* | | | | | | | | | | | | | | | |
| Classifier Confidence | 0.51 | **0.43** | **0.35** | 0.66 | 0.57 | 0.46 | 0.59 | 0.52 | 0.40 | 0.39 | 0.32 | 0.25 | 0.41 | 0.38 | 0.30 |
| KL Divergence | 0.59 | 0.31 | 0.24 | 0.66 | 0.66 | 0.54 | **0.62** | 0.62 | 0.50 | **0.46** | 0.46 | 0.36 | 0.51 | **0.60** | 0.49 |
| Cosine Similarity | -0.55 | -0.44 | -0.36 | -0.66 | -0.67 | -0.54 | -0.53 | -0.59 | -0.46 | -0.43 | -0.40 | -0.32 | -0.48 | -0.58 | -0.47 |
| Jensen-Shannon Divergence | **0.67** | 0.40 | 0.32 | **0.69** | **0.67** | **0.55** | **0.62** | **0.64** | **0.51** | 0.41 | **0.50** | **0.39** | 0.53 | **0.60** | **0.50** |
| Hybrid-Simulation | **0.69** | **0.40** | **0.32** | **0.71** | **0.67** | 0.54 | **0.62** | **0.64** | **0.51** | **0.44** | **0.47** | **0.37** | 0.53 | **0.61** | 0.49 |
| Hybrid-Learned | 0.67 | 0.37 | 0.30 | 0.70 | 0.63 | 0.50 | 0.61 | 0.62 | 0.49 | 0.43 | 0.47 | 0.37 | **0.55** | **0.62** | **0.50** |

Table 1: Style transfer quality (reference-free). Pearson (PC), Spearman (SC) and Kendall's Tau (KC) correlations.

|  | Sentiment Transfer | | | | | | | | | Detoxification | | | | | |
|---|---|---|---|---|---|---|---|---|---|---|---|---|---|---|---|
|  | English | | | Hindi | | | Bengali | | | English | | | Hindi | | |
| Metrics | PC | SC | KC | PC | SC | KC | PC | SC | KC | PC | SC | KC | PC | SC | KC |
| *Previously used & LLMs* | | | | | | | | | | | | | | | |
| BLEU | 0.24 | 0.22 | 0.18 | 0.24 | 0.19 | 0.15 | 0.32 | 0.31 | 0.25 | 0.14 | 0.13 | 0.11 | 0.45 | 0.37 | 0.31 |
| Cosine Similarity | **0.54** | **0.27** | **0.22** | 0.33 | 0.24 | 0.20 | 0.43 | 0.40 | 0.32 | 0.28 | 0.19 | 0.15 | 0.59 | 0.45 | 0.38 |
| Masked BLEU | 0.21 | 0.21 | 0.17 | 0.15 | 0.12 | 0.10 | 0.23 | 0.24 | 0.19 | 0.15 | 0.15 | 0.12 | 0.45 | 0.39 | 0.32 |
| Masked Cosine Similarity | 0.36 | 0.17 | 0.14 | 0.19 | 0.13 | 0.11 | 0.28 | 0.29 | 0.23 | 0.23 | 0.15 | 0.12 | 0.56 | **0.45** | 0.37 |
| METEOR | 0.38 | 0.25 | 0.21 | 0.20 | 0.18 | 0.14 | 0.33 | 0.27 | 0.22 | 0.16 | 0.10 | 0.08 | 0.54 | 0.34 | 0.28 |
| ROUGE-2 | 0.24 | 0.19 | 0.16 | 0.19 | 0.20 | 0.16 | 0.28 | 0.30 | 0.24 | 0.17 | 0.11 | 0.09 | 0.41 | 0.37 | 0.31 |
| ROUGE-L | 0.39 | 0.25 | 0.21 | 0.26 | 0.23 | 0.19 | 0.34 | 0.32 | 0.25 | 0.22 | 0.12 | 0.10 | 0.46 | 0.39 | 0.33 |
| GPT4 | **0.42** | **0.36** | **0.35** | **0.39** | **0.41** | **0.39** | **0.51** | **0.54** | **0.48** | **0.46** | **0.31** | **0.30** | **0.46** | **0.42** | **0.40** |
| Llama | 0.24 | 0.26 | 0.24 | 0.32 | 0.28 | 0.26 | 0.32 | 0.38 | 0.35 | 0.25 | 0.11 | 0.10 | 0.28 | 0.16 | 0.16 |
| *Newly applied & Novel* | | | | | | | | | | | | | | | |
| PINC | -0.18 | -0.17 | -0.15 | -0.16 | -0.12 | -0.10 | -0.27 | -0.28 | -0.23 | -0.12 | -0.12 | -0.10 | -0.41 | -0.36 | -0.30 |
| WMD | 0.35 | 0.28 | 0.23 | 0.27 | 0.24 | 0.20 | 0.34 | 0.35 | 0.28 | 0.15 | 0.14 | 0.11 | 0.41 | 0.38 | 0.32 |
| BERTScore | **0.50** | **0.31** | **0.26** | 0.45 | 0.33 | 0.27 | 0.49 | 0.44 | 0.36 | 0.21 | 0.19 | 0.15 | **0.62** | 0.38 | 0.31 |
| Smatch (Dependency Trees) | 0.25 | 0.24 | 0.20 | 0.18 | 0.20 | 0.17 | 0.26 | 0.30 | 0.25 | 0.16 | 0.15 | 0.12 | 0.34 | 0.31 | 0.26 |
| Smatch (AMR) | 0.38 | 0.25 | 0.20 | 0.22 | 0.20 | 0.17 | 0.32 | 0.32 | 0.26 | 0.19 | 0.13 | 0.11 | 0.37 | 0.34 | 0.28 |
| S3BERT | 0.46 | 0.23 | 0.19 | 0.24 | 0.30 | 0.18 | 0.14 | 0.30 | 0.30 | 0.22 | 0.20 | 0.16 | 0.49 | 0.38 | 0.31 |
| BLEURT | 0.47 | 0.30 | 0.25 | 0.41 | **0.35** | **0.29** | 0.56 | **0.53** | **0.42** | 0.18 | 0.17 | 0.14 | **0.62** | **0.43** | **0.35** |
| TER | 0.42 | 0.26 | 0.22 | **0.45** | 0.28 | 0.24 | 0.34 | 0.33 | 0.27 | 0.21 | 0.17 | 0.14 | 0.58 | 0.37 | 0.31 |
| TED | 0.43 | 0.24 | 0.22 | 0.42 | 0.29 | 0.25 | 0.20 | 0.28 | 0.24 | **0.48** | **0.21** | **0.18** | 0.48 | 0.36 | 0.30 |
| Hybrid-Simulation | **0.57** | **0.32** | **0.26** | **0.48** | 0.33 | 0.27 | **0.57** | **0.53** | **0.43** | 0.28 | 0.19 | 0.15 | **0.68** | **0.43** | **0.35** |
| Hybrid-Learned | 0.56 | **0.32** | **0.26** | 0.47 | **0.35** | **0.29** | 0.56 | **0.53** | **0.43** | 0.19 | 0.15 | 0.12 | 0.64 | 0.38 | 0.31 |

Table 2: Content preservation (reference-free). Pearson (PC), Spearman (SC) and Kendall's Tau (KC) correlations.

|  | Sentiment Transfer | | | | | | | | | Detoxification | | | | | |
|---|---|---|---|---|---|---|---|---|---|---|---|---|---|---|---|
|  | English | | | Hindi | | | Bengali | | | English | | | Hindi | | |
| Metrics | PC | SC | KC | PC | SC | KC | PC | SC | KC | PC | SC | KC | PC | SC | KC |
| *Previously used & LLMs* | | | | | | | | | | | | | | | |
| Perplexity (GPT-2) | **0.13** | 0.13 | 0.11 | -0.11 | -0.10 | -0.08 | -0.11 | -0.07 | -0.05 | **0.06** | **0.00** | **0.00** | 0.17 | -0.13 | -0.11 |
| Perplexity (MGPT) | 0.08 | **0.19** | **0.15** | **0.00** | **0.07** | **0.05** | 0.16 | 0.19 | 0.15 | 0.05 | **0.00** | **0.00** | 0.11 | **0.03** | **0.03** |
| GPT4 | **0.43** | **0.40** | **0.37** | **0.39** | **0.39** | **0.35** | **0.37** | **0.40** | **0.36** | 0.16 | 0.13 | 0.12 | 0.17 | 0.17 | 0.16 |
| Llama | 0.17 | 0.18 | 0.17 | 0.15 | 0.17 | 0.15 | 0.08 | 0.06 | 0.06 | **0.16** | **0.13** | **0.12** | -0.01 | -0.02 | -0.01 |
| *Newly applied* | | | | | | | | | | | | | | | |
| Perplexity (Finetuned GPT-2) | **0.14** | **0.16** | **0.13** | 0.08 | 0.14 | 0.11 | 0.02 | 0.05 | 0.04 | **0.14** | 0.00 | 0.00 | 0.11 | -0.06 | -0.05 |
| Perplexity (Finetuned MGPT) | 0.04 | 0.08 | 0.07 | **0.17** | **0.15** | **0.12** | **0.23** | **0.21** | **0.16** | 0.00 | **0.03** | **0.03** | **0.23** | **0.04** | **0.03** |

Table 3: Fluency (reference-free). Pearson (PC), Spearman (SC) and Kendall's Tau (KC) correlations.

when the stylistic shift involves changing sentiment. However, for detoxification tasks, its alignment with human judgments diminishes, indicating that removing toxicity poses different challenges for GPT-4. In contrast, *Llama* exhibits generally weaker correlations and struggles in various settings, implying that its evaluations of fluency do not consistently match human perceptions.

Language-wise, English generally shows better correlations and less variability across models over

|  | Sentiment Transfer | | | | | | | | | Detoxification | | | | | |
|---|---|---|---|---|---|---|---|---|---|---|---|---|---|---|---|
|  | English | | | Hindi | | | Bengali | | | English | | | Hindi | | |
| Metrics | PC | SC | KC | PC | SC | KC | PC | SC | KC | PC | SC | KC | PC | SC | KC |
| Existing | 0.32 | 0.02 | 0.02 | 0.11 | -0.02 | -0.01 | 0.25 | 0.18 | 0.13 | -0.04 | -0.18 | -0.14 | 0.07 | -0.19 | -0.14 |
| GPT4 | **0.73** | **0.62** | **0.54** | **0.78** | **0.75** | **0.61** | **0.78** | **0.77** | **0.63** | **0.65** | **0.62** | **0.51** | 0.62 | **0.59** | **0.46** |
| Llama | 0.08 | 0.16 | 0.13 | 0.02 | 0.01 | 0.01 | 0.01 | 0.01 | 0.00 | 0.18 | 0.14 | 0.11 | 0.27 | 0.23 | 0.19 |
| Ours$_1$ | 0.57 | 0.33 | 0.26 | 0.59 | 0.54 | 0.43 | 0.54 | 0.57 | 0.42 | 0.38 | 0.44 | 0.34 | 0.47 | 0.43 | 0.32 |
| Ours$_2$ | 0.68 | 0.40 | 0.31 | 0.72 | 0.68 | 0.53 | 0.59 | 0.59 | 0.42 | 0.41 | 0.38 | 0.29 | **0.63** | 0.57 | 0.43 |

Table 4: Overall results (reference-free). Pearson (PC), Spearman (SC) and Kendall's Tau (KC) correlations.

Hindi and Bengali results.

### 5.4 Overall Score

Table 4 shows results for the different versions of the overall score aggregating style transfer accuracy, content preservation, and fluency.

**Previously Used:** Aggregating traditional metrics in the *Existing* metric often yields near-zero or negative correlations across various languages and tasks, indicating that simply merging these measures fails to capture the overall quality.

**LLMs:** In contrast, *GPT-4* consistently aligns well with human assessments of overall quality in both Sentiment Transfer and Detoxification. *Llama*, however, shows weaker correlations, indicating that not all LLMs possess the same evaluative capabilities.

**Newly Applied & Hybrid:** Our approaches (*Ours$_1$* and *Ours$_2$*) provide noticeable improvements over existing methods. Although they do not surpass GPT-4, they clearly outperform many traditional and alternative measures.

## 6 Conclusion

We presented a comprehensive evaluation of existing and newly proposed metrics for two TST subtasks—*Sentiment Transfer* and *Text Detoxification*—in English, Hindi, and Bengali. Our findings demonstrate that traditional word-overlap-based metrics like BLEU and ROUGE often show limited correlation with human judgments, whereas our proposed experimental metrics and prompted LLM-based evaluations provide significantly stronger alignment. Moreover, our oracle hybrid ensemble and combined approaches show an even greater potential of merging multiple metrics.

## Limitations

Our study is limited to two specific tasks and three languages, leaving open the question of how well these metrics generalize to other styles, languages, and domains as future work. Additionally, while oracle ensemble metrics provide valuable insights, further research is needed to develop fully generalizable evaluation methods that do not rely on target-specific tuning.

## Acknowledgments

This research was funded by the European Union (ERC, NG-NLG, 101039303) and Charles University project SVV 260 698. We acknowledge the use of resources provided by the LINDAT/CLARIAH-CZ Research Infrastructure (Czech Ministry of Education, Youth, and Sports project No. LM2018101). We also acknowledge Panlingua Language Processing LLP for collaborating on this research project.
Atul Kr. Ojha and John P. McCrae would like to acknowledge the support of the Research Ireland as part of Grant Number SFI/12/RC/2289_P2 Insight_2, Insight Research Ireland Centre for Data Analytics.

# A Hybrid Approaches and Overall Score - Additional Details

In this section, we introduce our hybrid approaches by presenting both the selected metrics and their associated simulated weights, as well as the learned normalized feature importance. Further details on these weights, selected metrics, and feature scores can be found in Tables 5 and 6 respectively. Table 7 summarizes the selected metrics for each language and task, enabling single overall scores computation.

|  | Sentiment Transfer | | | | | | Detoxification | | | |
|---|---|---|---|---|---|---|---|---|---|---|
|  | Simulation | | | Learned | | | Simulation | | Learned | |
| Metrics | English | Hindi | Bengali | English | Hindi | Bengali | English | Hindi | English | Hindi |
| BERTScore | 0.20 | 0.40 | 0.40 | - | 0.36 | 0.14 | - | - | - | 0.30 |
| BERTScore_IDF | - | - | - | 0.27 | 0.35 | - | - | - | - | 0.11 |
| BLEURT | 0.30 | 0.20 | 0.50 | 0.43 | 0.29 | 0.65 | - | 0.40 | - | - |
| BLEU | - | - | - | - | - | - | - | - | 0.34 | - |
| Masked BLEU | - | - | - | - | - | - | - | - | 0.25 | - |
| COSINE | 0.50 | - | 0.10 | 0.30 | - | 0.21 | 0.20 | 0.30 | - | - |
| TER | - | 0.40 | - | - | - | - | 0.10 | 0.30 | - | 0.59 |
| TED | - | - | - | - | - | - | 0.70 | - | 0.40 | - |

|  | Sentiment Transfer | | | | | | Detoxification | | | |
|---|---|---|---|---|---|---|---|---|---|---|
|  | Simulation | | | Learned | | | Simulation | | Learned | |
| Metrics | English | Hindi | Bengali | English | Hindi | Bengali | English | Hindi | English | Hindi |
| EMD | - | - | - | - | 0.33 | 0.24 | - | - | - | - |
| JS | 0.60 | 0.40 | 0.40 | 0.38 | 0.46 | 0.35 | 0.30 | 0.30 | 0.42 | 0.45 |
| KL | 0.15 | 0.20 | 0.30 | 0.38 | - | 0.41 | 0.50 | 0.50 | 0.27 | 0.22 |
| Style_Classifier_Confidence | 0.25 | 0.40 | 0.30 | 0.24 | 0.21 | - | 0.20 | 0.20 | 0.31 | 0.33 |

Table 5: Hybrid Simulation - selected metrics and its weights.

|  | Sentiment Transfer (CS) | | | Detoxification | |
|---|---|---|---|---|---|
| Metrics | English | Hindi | Bengali | English | Hindi |
| BLEURT | 0.16 | 0.13 | 0.37 | 0.05 | 0.04 |
| COSINE | 0.11 | 0.08 | 0.12 | 0.08 | 0.04 |
| BERTScore_IDF | 0.10 | 0.16 | 0.03 | 0.05 | 0.19 |
| BERTScore | 0.09 | 0.17 | 0.08 | 0.05 | 0.07 |
| S3BERT | 0.07 | 0.08 | 0.04 | 0.05 | 0.02 |
| WMD | 0.07 | 0.03 | 0.03 | 0.04 | 0.01 |
| AMR_SMATCH | 0.06 | 0.02 | 0.02 | 0.05 | 0.02 |
| BLEU | 0.06 | 0.03 | 0.07 | 0.12 | 0.03 |
| ROUGE-L | 0.06 | 0.02 | 0.03 | 0.07 | 0.05 |
| Masked Cosine Similarity | 0.06 | 0.02 | 0.02 | 0.06 | 0.04 |
| Masked BLEU | 0.05 | 0.04 | 0.03 | 0.09 | 0.02 |
| METEOR | 0.03 | 0.04 | 0.04 | 0.05 | 0.02 |
| TED | 0.02 | 0.04 | 0.02 | 0.14 | 0.02 |
| SMATCH | 0.02 | 0.02 | 0.02 | 0.02 | 0.03 |
| TER | 0.02 | 0.16 | 0.04 | 0.03 | 0.38 |
| ROUGE-2 | 0.01 | 0.02 | 0.03 | 0.04 | 0.01 |
| PINC | 0.01 | 0.01 | 0.01 | 0.02 | 0.01 |

|  | Sentiment Transfer (SA) | | | Detoxification | |
|---|---|---|---|---|---|
| Metrics | English | Hindi | Bengali | English | Hindi |
| KL | 0.34 | 0.17 | 0.33 | 0.21 | 0.18 |
| JS | 0.33 | 0.38 | 0.29 | 0.33 | 0.37 |
| Style_Classifier_Confidence | 0.21 | 0.17 | 0.18 | 0.25 | 0.26 |
| EMD | 0.11 | 0.27 | 0.20 | 0.21 | 0.18 |
| Sentence_Accuracy | 0.01 | 0.00 | 0.00 | 0.00 | 0.01 |

Table 6: Hybrid-Learned - metrics and its learned feature importance scores (normalized).

# B Additional Results (reference-based)

In addition to the reference-free evaluations shown in Tables 1 and 2, the corresponding reference-based results are provided in Tables 8 and 9, respectively.

| Task | Languages | Approach | BERTScore | BLEURT | Cosine Similarity | Hybrid_Learned_CP | Hybrid_Simulation_CP | Hybrid_Simulation_ST | JS | KL | Perplexity (MGPT) | MGPT_FT_PPL | Perplexity (GPT-2) | GPT2_FT_PPL | Sentence Accuracy | TED | TER |
|---|---|---|---|---|---|---|---|---|---|---|---|---|---|---|---|---|---|
| Sentiment Transfer | English | Existing | ✗ | ✗ | ✓ | ✗ | ✗ | ✗ | ✗ | ✗ | ✗ | ✓ | ✗ | ✓ | ✗ | ✗ |
| | | Ours₁ | ✓ | ✗ | ✗ | ✗ | ✗ | ✗ | ✓ | ✗ | ✗ | ✗ | ✗ | ✓ | ✗ | ✗ |
| | | Ours₂ | ✗ | ✗ | ✗ | ✗ | ✓ | ✓ | ✗ | ✗ | ✗ | ✗ | ✗ | ✗ | ✗ | ✗ |
| | Hindi | Existing | ✗ | ✗ | ✓ | ✗ | ✗ | ✗ | ✗ | ✗ | ✓ | ✗ | ✗ | ✗ | ✓ | ✗ |
| | | Ours₁ | ✗ | ✗ | ✗ | ✗ | ✗ | ✗ | ✓ | ✗ | ✗ | ✓ | ✗ | ✗ | ✗ | ✓ |
| | | Ours₂ | ✗ | ✗ | ✗ | ✗ | ✓ | ✓ | ✗ | ✗ | ✗ | ✗ | ✗ | ✗ | ✗ | ✗ |
| | Bengali | Existing | ✗ | ✗ | ✓ | ✗ | ✗ | ✗ | ✗ | ✗ | ✗ | ✓ | ✗ | ✓ | ✗ | ✗ |
| | | Ours₁ | ✗ | ✓ | ✗ | ✗ | ✗ | ✗ | ✓ | ✗ | ✗ | ✗ | ✗ | ✗ | ✗ | ✗ |
| | | Ours₂ | ✗ | ✗ | ✗ | ✗ | ✓ | ✓ | ✗ | ✗ | ✗ | ✗ | ✓ | ✗ | ✗ | ✗ |
| Detoxification | English | Existing | ✗ | ✗ | ✓ | ✗ | ✗ | ✗ | ✗ | ✗ | ✗ | ✓ | ✗ | ✓ | ✗ | ✗ |
| | | Ours₁ | ✗ | ✗ | ✗ | ✗ | ✗ | ✗ | ✗ | ✓ | ✗ | ✗ | ✗ | ✓ | ✓ | ✗ |
| | | Ours₂ | ✗ | ✗ | ✗ | ✓ | ✓ | ✗ | ✗ | ✗ | ✗ | ✗ | ✗ | ✓ | ✗ | ✗ |
| | Hindi | Existing | ✗ | ✗ | ✓ | ✗ | ✗ | ✗ | ✗ | ✓ | ✗ | ✗ | ✗ | ✓ | ✗ | ✗ |
| | | Ours₁ | ✗ | ✓ | ✗ | ✗ | ✗ | ✗ | ✓ | ✗ | ✗ | ✗ | ✗ | ✗ | ✗ | ✗ |
| | | Ours₂ | ✗ | ✗ | ✗ | ✓ | ✓ | ✗ | ✗ | ✓ | ✗ | ✗ | ✗ | ✗ | ✗ | ✗ |

Table 7: Overall Scores – language and task-wise selected metrics.

| | Sentiment Transfer (reference-based) | | | | | | | | | Detoxification (reference-based) | | | | | |
|---|---|---|---|---|---|---|---|---|---|---|---|---|---|---|---|
| | English | | | Hindi | | | Bengali | | | English | | | Hindi | | |
| Metrics | PC | SC | KC | PC | SC | KC | PC | SC | KC | PC | SC | KC | PC | SC | KC |
| EMD | -0.22 | -0.27 | -0.22 | -0.28 | -0.33 | -0.26 | -0.33 | -0.37 | -0.29 | -0.28 | -0.23 | -0.18 | -0.31 | -0.28 | -0.22 |
| KL_DIS | -0.30 | -0.36 | -0.29 | -0.62 | -0.58 | -0.46 | -0.46 | -0.48 | -0.38 | -0.34 | -0.30 | -0.24 | -0.36 | -0.28 | -0.23 |
| Cosine Similarity | **0.32** | **0.32** | **0.26** | **0.59** | **0.60** | **0.49** | **0.34** | **0.45** | **0.35** | **0.28** | **0.32** | **0.25** | **0.30** | **0.32** | **0.26** |
| JS_SIM | -0.29 | -0.36 | -0.29 | -0.62 | -0.58 | -0.46 | -0.46 | -0.47 | -0.37 | -0.28 | -0.29 | -0.23 | -0.35 | -0.28 | -0.23 |

Table 8: Automatic metrics results reference-based: style transfer. Pearson Correlation: PC, Spearman Correlation: SC Kendall Tau Correlation: KC

| | Sentiment Transfer (reference-based) | | | | | | | | | Detoxification (reference-based) | | | | | |
|---|---|---|---|---|---|---|---|---|---|---|---|---|---|---|---|
| | English | | | Hindi | | | Bengali | | | English | | | Hindi | | |
| Metrics | PC | SC | KC | PC | SC | KC | PC | SC | KC | PC | SC | KC | PC | SC | KC |
| *Previously used & LLMs* | | | | | | | | | | | | | | | |
| BLEU | 0.18 | 0.22 | 0.18 | 0.17 | 0.16 | 0.13 | 0.19 | 0.20 | 0.16 | 0.10 | 0.10 | 0.08 | 0.18 | 0.18 | 0.14 |
| Cosine Similarity | 0.39 | 0.26 | 0.22 | 0.20 | 0.24 | 0.19 | 0.31 | 0.32 | 0.25 | **0.18** | 0.13 | 0.10 | 0.30 | 0.25 | 0.20 |
| Masked BLEU | 0.13 | 0.19 | 0.15 | 0.16 | 0.16 | 0.13 | 0.18 | 0.17 | 0.13 | 0.11 | 0.11 | 0.09 | 0.16 | 0.15 | 0.12 |
| Masked Cosine Similarity | 0.25 | 0.21 | 0.17 | 0.15 | 0.16 | 0.13 | 0.24 | 0.28 | 0.22 | 0.17 | **0.13** | **0.10** | 0.24 | 0.22 | 0.18 |
| METEOR | 0.31 | 0.22 | 0.18 | 0.12 | 0.13 | 0.10 | 0.16 | 0.18 | 0.14 | 0.11 | 0.09 | 0.08 | 0.23 | 0.17 | 0.14 |
| ROUGE-2 | 0.22 | 0.21 | 0.17 | 0.17 | 0.18 | 0.15 | 0.23 | 0.23 | 0.18 | 0.11 | 0.09 | 0.07 | 0.24 | 0.24 | 0.20 |
| ROUGE-L | 0.31 | 0.24 | 0.20 | 0.19 | 0.19 | 0.16 | 0.21 | 0.23 | 0.18 | 0.13 | 0.09 | 0.07 | 0.23 | 0.24 | 0.19 |
| *Newly applied & Novel* | | | | | | | | | | | | | | | |
| PINC | -0.12 | -0.14 | -0.12 | -0.13 | -0.12 | -0.10 | -0.17 | -0.18 | -0.16 | -0.09 | -0.07 | -0.06 | -0.17 | -0.15 | -0.13 |
| WMD | 0.25 | 0.26 | 0.21 | 0.21 | 0.22 | 0.18 | 0.25 | 0.27 | 0.21 | 0.11 | 0.08 | 0.07 | 0.19 | 0.19 | 0.15 |
| BERTScore | 0.34 | **0.27** | **0.22** | 0.25 | 0.25 | 0.20 | 0.24 | 0.25 | 0.20 | 0.18 | 0.16 | **0.13** | 0.32 | 0.19 | 0.15 |
| UDPIPE_SMATCH | 0.16 | 0.20 | 0.16 | 0.19 | 0.19 | 0.16 | 0.18 | 0.18 | 0.14 | 0.16 | 0.15 | 0.12 | 0.15 | 0.14 | 0.12 |
| AMR_SMATCH | 0.28 | **0.27** | 0.22 | 0.22 | 0.20 | 0.17 | 0.25 | 0.24 | 0.19 | 0.12 | 0.09 | 0.07 | 0.18 | 0.17 | 0.14 |
| S3BERT | **0.41** | 0.26 | 0.21 | 0.28 | 0.22 | 0.18 | 0.23 | 0.26 | 0.21 | 0.13 | 0.13 | 0.11 | 0.26 | 0.20 | 0.16 |
| BLEURT | 0.31 | 0.25 | 0.20 | **0.31** | **0.31** | **0.25** | **0.42** | **0.41** | **0.32** | 0.15 | **0.17** | **0.13** | **0.35** | **0.23** | **0.19** |
| TER | 0.35 | 0.23 | 0.19 | **0.39** | 0.26 | 0.21 | 0.22 | 0.21 | 0.17 | **0.24** | 0.10 | 0.09 | 0.23 | 0.14 | 0.12 |
| TED | -0.29 | -0.23 | -0.20 | -0.35 | -0.26 | -0.22 | -0.17 | -0.15 | -0.13 | -0.40 | -0.16 | -0.13 | -0.32 | -0.20 | -0.17 |

Table 9: Automatic metrics results reference-based: content preservation. Pearson Correlation: PC, Spearman Correlation: SC Kendall Tau Correlation: KC

## C GPT-2 and MGPT Finetune Details

We fine-tune both *GPT-2*[3] and *mGPT*[4] using the same hyperparameter configuration obtained through few random optimization experiments. Specifically, we set the maximum token length to 512 and use the target-style training data from (Mukherjee et al., 2024b) for fine-tuning. Each model is trained for 10 epochs with a batch size of 2, a learning rate of $1 \times 10^{-5}$, and a weight decay of 0.01.

## D Prompt Details

This section provides a collection of example prompts (in English) for the evaluation of Text Sentiment Transfer task (prompt details in Tables 10, 11, 12 and 13)

---
[3] https://huggingface.co/openai-community/gpt2
[4] https://huggingface.co/ai-forever/mGPT

| | |
|---|---|
| **Prompt** | Sentiment transfer task: transfer the sentiment of a sentence (from positive to negative or negative to positive) while keeping the rest of the sentiment-independent content unchanged.<br><br>Please rate the sentiment transfer accuracy of the negative to positive sentiment transfer task between the following English source sentence S1 and the sentiment-transferred sentence S2. Use a scale of 1 to 5, where 1 indicates that the sentiment in S1 is completely identical to the sentiment in S2, and 5 indicates that the sentiment has been completely transferred to the target sentiment in S2.<br><br>S1: so he can charge a bloody fortune for them.<br>S2: so he can charge a fair amount of money for them.<br><br>Sentiment transfer accuracy rating (on a scale of 1 to 5) = |

Table 10: A Sample prompt for Sentiment Transfer Accuracy evaluation in Sentiment Transfer in English. It contains task definition, instruction, and input.

| | |
|---|---|
| **Prompt** | Sentiment transfer task: transfer the sentiment of a sentence (from positive to negative or negative to positive) while keeping the rest of the sentiment-independent content unchanged.<br><br>Please act as a binary classifier to evaluate the sentiment transfer accuracy of the positive to negative sentiment transfer task in English. Determine whether the target sentiment has been successfully transferred to the generated sentence (S2) from the source sentence (S1). If the target sentiment has been successfully transferred to S2, output '1'. If the target sentiment has not been successfully transferred to S2, output '0'.<br><br>S1: so he can charge a bloody fortune for them.<br>S2: so he can charge a fair amount of money for them.<br><br>Sentiment transfer accuracy classification (0 or 1) = |

Table 11: A Sample prompt for Sentiment Transfer Accuracy (binary) evaluation in Sentiment Transfer in English. It contains task definition, instruction, and input.

| | |
|---|---|
| **Prompt** | Sentiment transfer task: transfer the sentiment of a sentence (from positive to negative or negative to positive) while keeping the rest of the content unchanged.<br><br>Please rate the content preservation between the following English source sentence S1 and the sentiment-transferred sentence S2 for the negative to positive sentiment transfer task on a scale of 1 to 5, where 1 indicates very low content preservation and 5 indicates very high content preservation. To determine the content preservation between these two sentences, consider only the information conveyed by the sentences and ignore any differences in sentiment due to the negative to positive sentiment transfer.<br><br>S1: so he can charge a bloody fortune for them.<br>S2: so he can charge a fair amount of money for them.<br><br>Content Preservation rating (on a scale of 1 to 5) = |

Table 12: A sample prompt for Content Preservation evaluation in Sentiment Transfer in English. It contains task definition, instruction, and input.

| | |
|---|---|
| **Prompt** | Please rate the fluency of the following English sentence S on a scale of 1 to 5, where 1 represents poor fluency, and 5 represents excellent fluency.<br><br>S: so he can charge a fair amount of money for them.<br><br>Fluency rating (on a scale of 1 to 5) = |

Table 13: A same prompt for Fluency evaluation in Sentiment Transfer in English. It contains instruction, and input.

# E Additional Statistics

In this section, we provide additional statistics for the Text Sentiment Transfer task in English, focusing on reference-free evaluation metrics. Specifically, we present heatmaps illustrating the correlations between each pair of metrics for style transfer accuracy, content preservation, and fluency in Figures 4, 5, and 6, respectively. We also show the distribution of each metric's values in Figures 1, 2, and 3 for style transfer accuracy, content preservation, and fluency, thereby offering a more comprehensive view of their behavior.

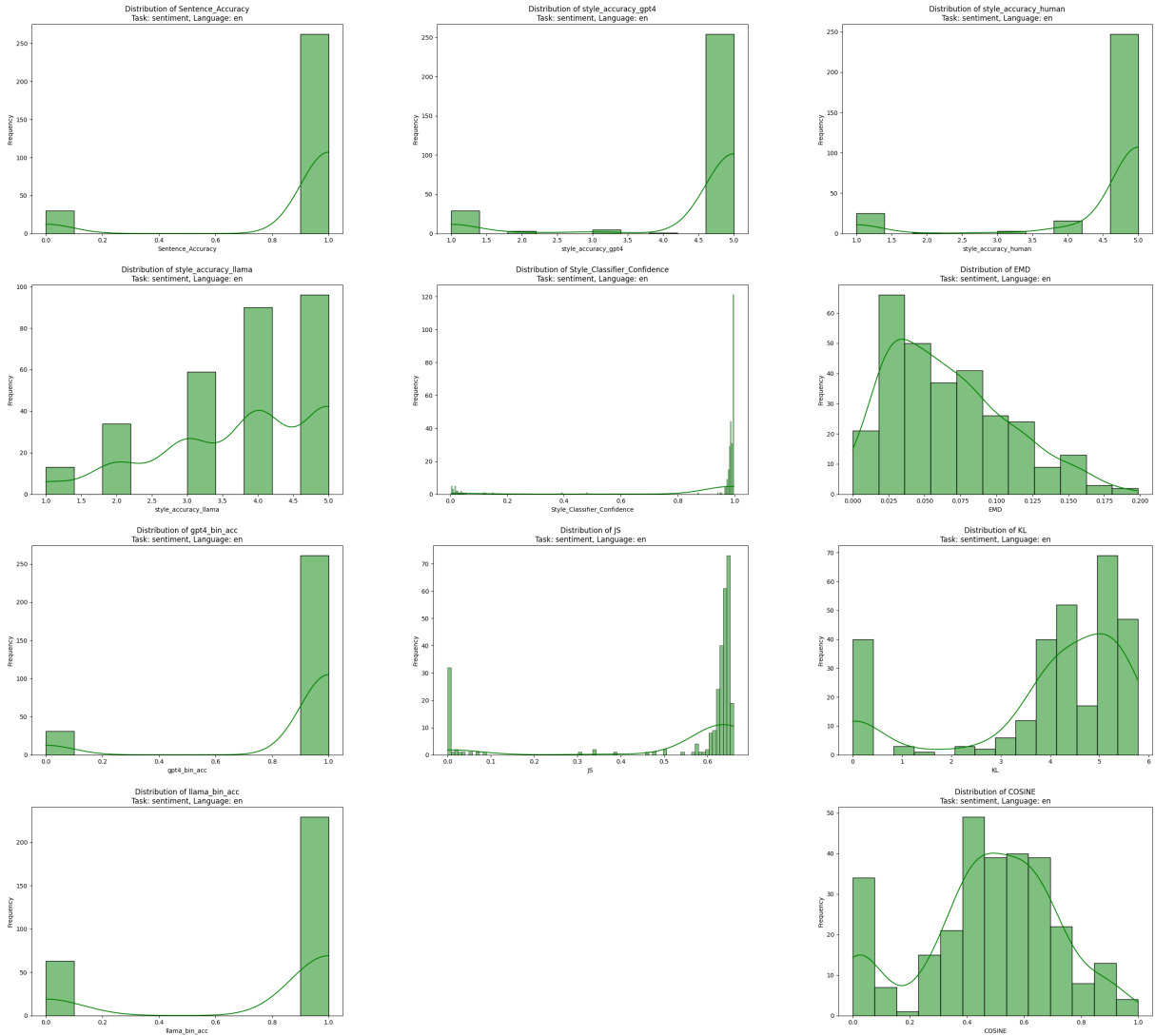

Figure 1: Style Transfer Accuracy - metrics' value distribution.

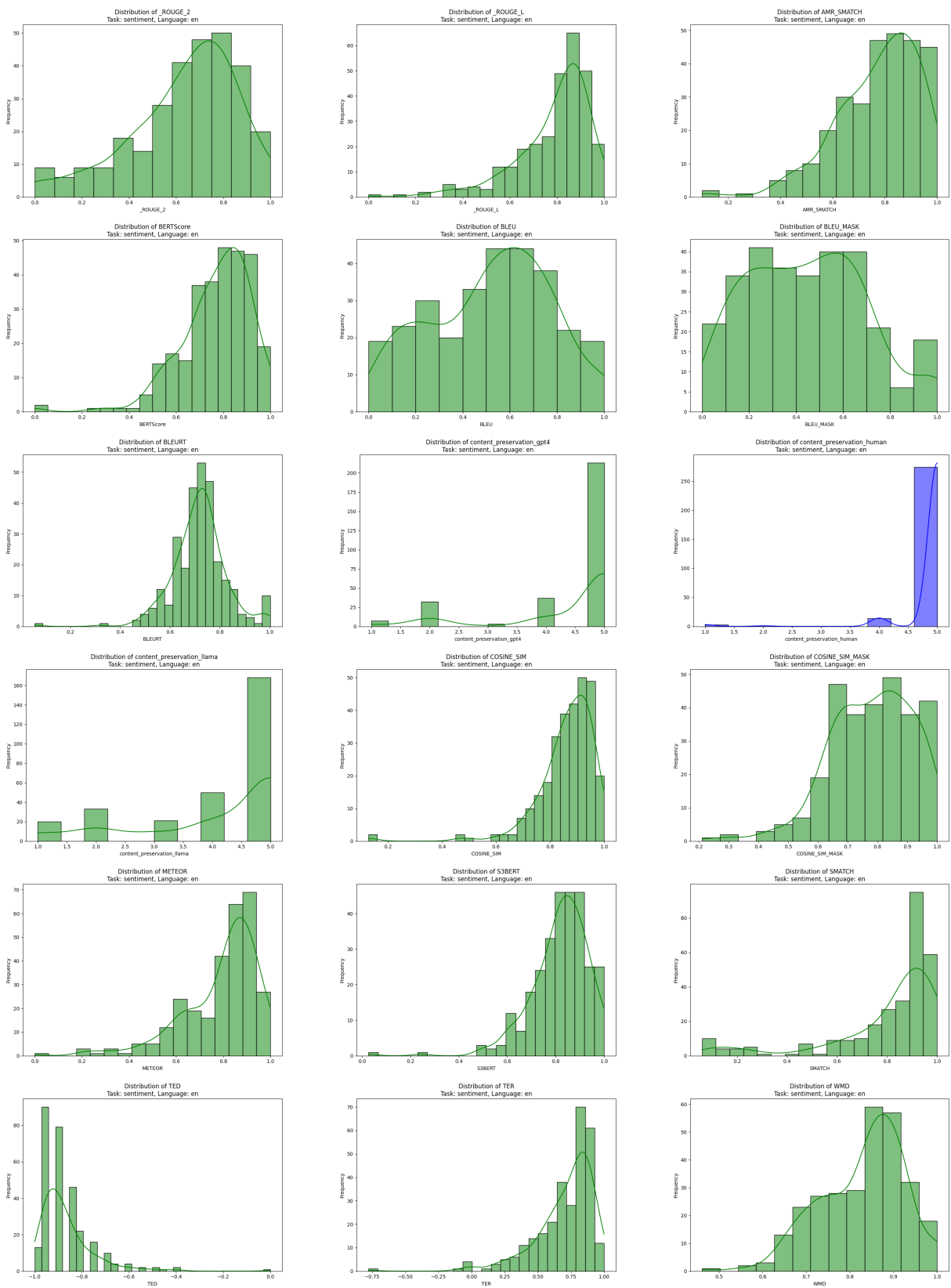

Figure 2: Content Preservation- - metrics' value distribution.

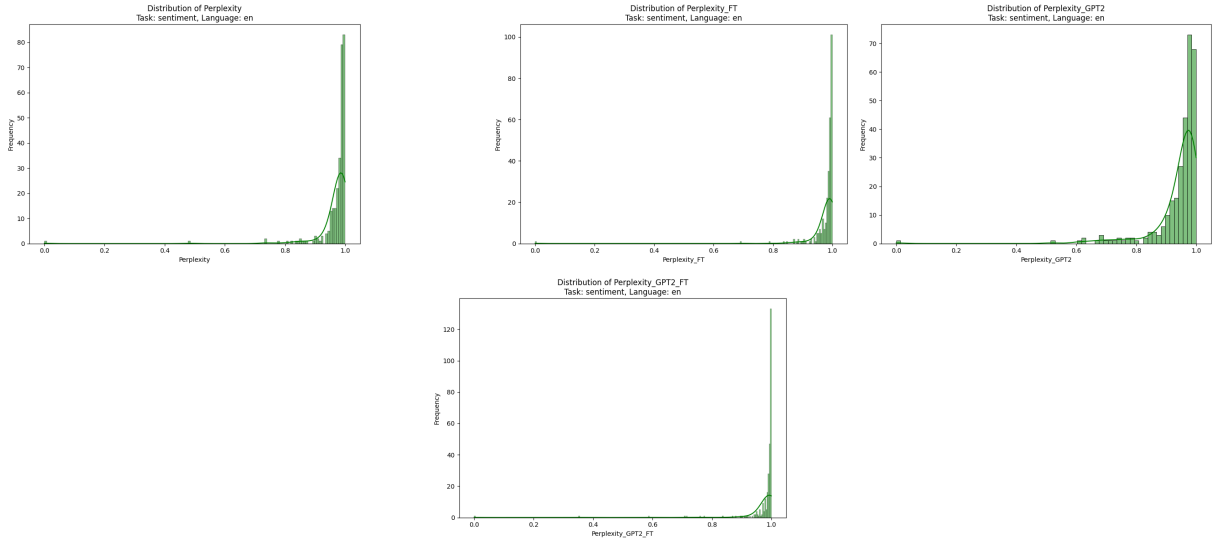

Figure 3: Fluency - metrics' value distribution.

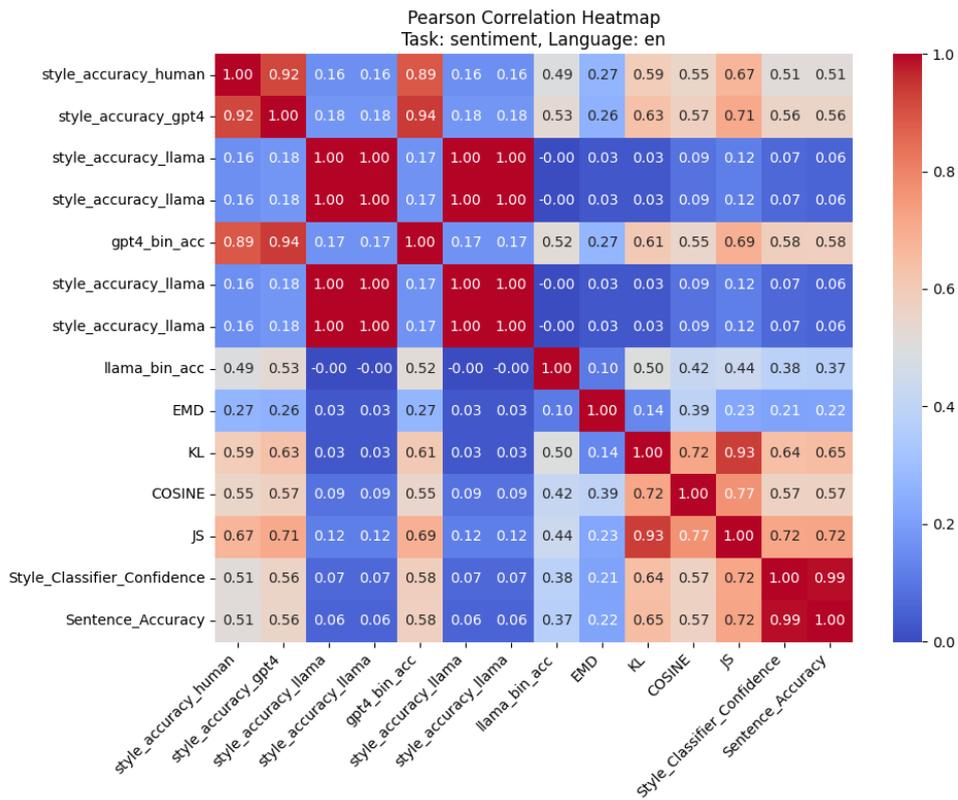

Figure 4: Sentence Accuracy - correlations' heatmap between the metrics.

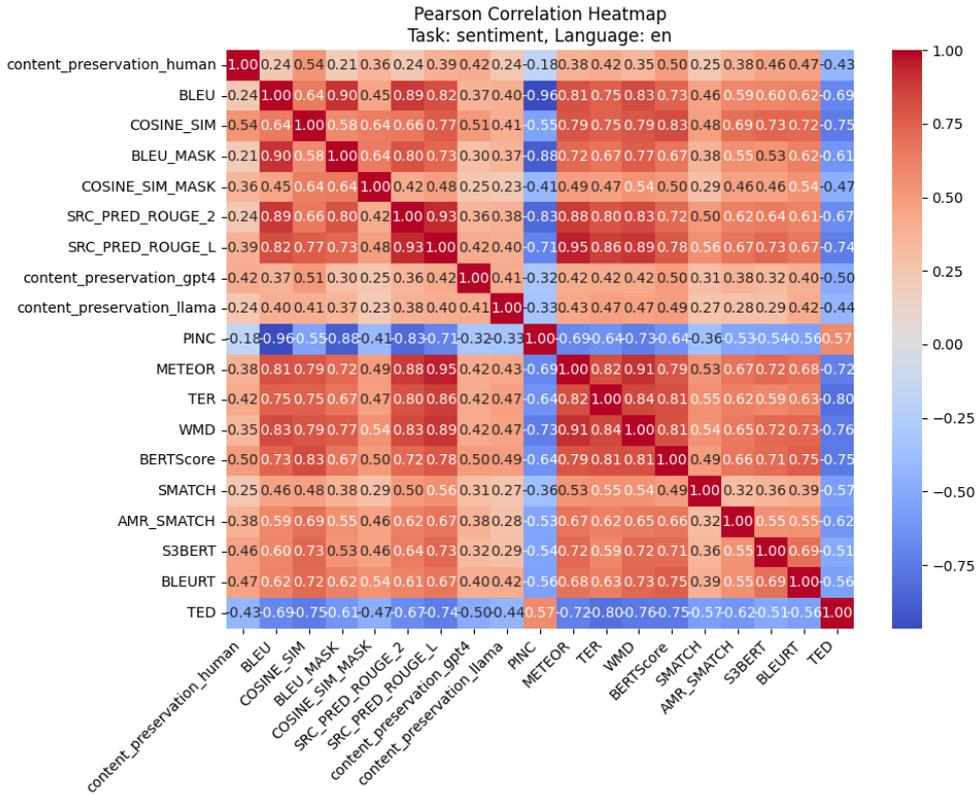

Figure 5: Content Preservation - correlations' heatmap between the metrics.

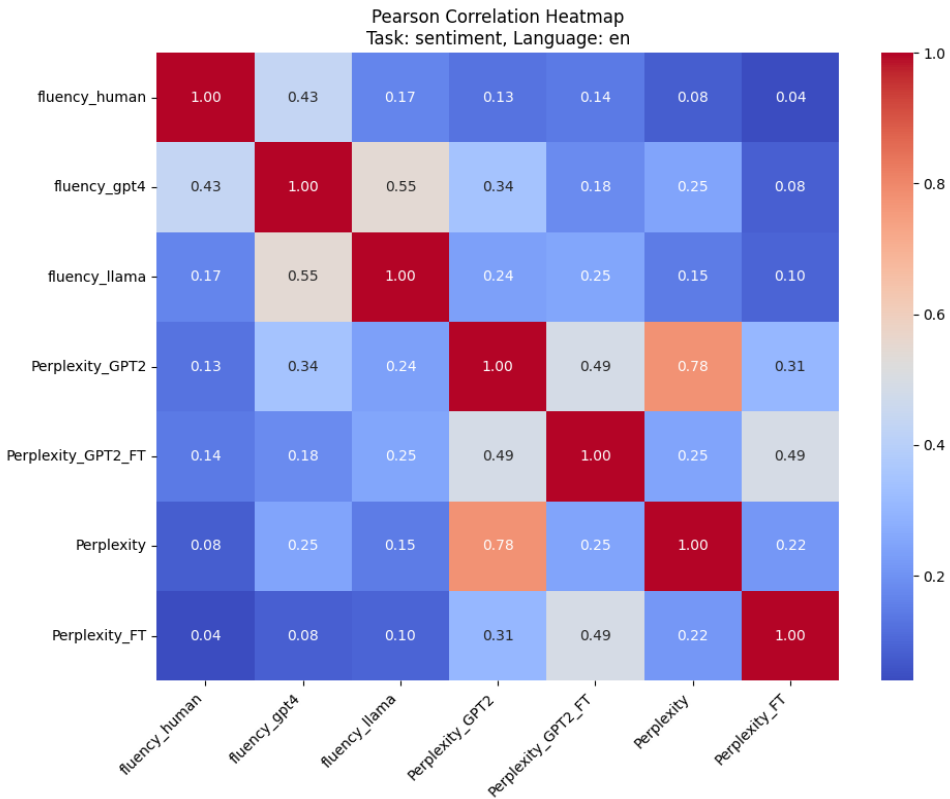

Figure 6: Fluency - correlations' heatmap between the metrics.